%% file: main.tex
\documentclass[10pt,twocolumn,letterpaper]{article}

\usepackage[pagenumbers]{cvpr} 

\usepackage[accsupp]{axessibility}

\input{preamble}

\definecolor{cvprblue}{rgb}{0.21,0.49,0.74}
\usepackage[pagebackref,breaklinks,colorlinks,citecolor=cvprblue]{hyperref}

\def\confName{CVPR}
\def\confYear{2024}

\title{FinePOSE: Fine-Grained Prompt-Driven 3D Human Pose Estimation via Diffusion Models}

\author{
Jinglin Xu\textsuperscript{1}\quad
Yijie Guo\textsuperscript{2}\quad
Yuxin Peng\textsuperscript{2}\thanks{}\\
\textsuperscript{1} School of Intelligence Science and Technology, University of Science and Technology Beijing\\
\textsuperscript{2} Wangxuan Institute of Computer Technology, Peking University\\
{\tt\small xujinglinlove@gmail.com; 2000012936@stu.pku.edu.cn; pengyuxin@pku.edu.cn}
}

\begin{document}
\twocolumn[{
\renewcommand\twocolumn[1][]{#1}
\maketitle
\begin{center}
\vspace{-12mm}
    \centering
    \captionsetup{type=figure}
    \includegraphics[width=\linewidth]{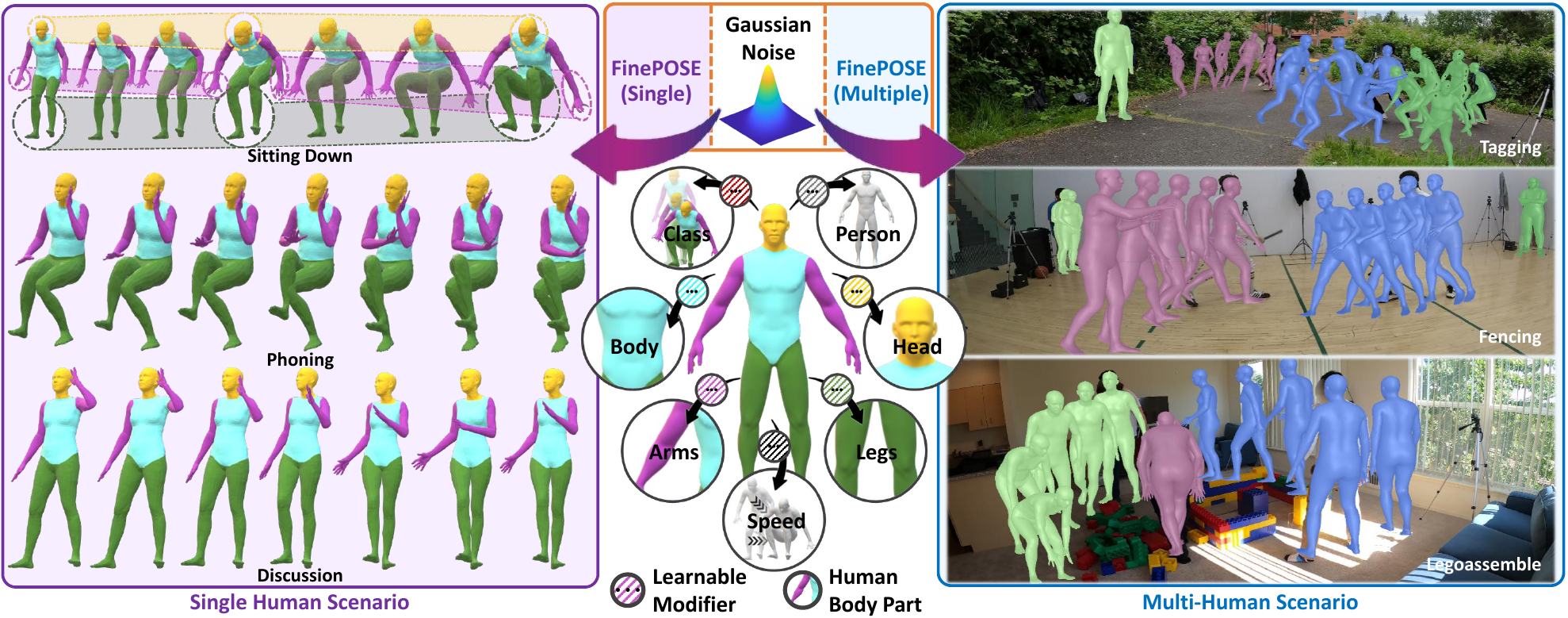}
    \vspace{-20pt}
    \caption{\textbf{Illustration of Fine-grained Prompt-driven Denoiser (FinePOSE).} FinePOSE, the proposed diffusion model-based 3D human pose estimation approach, enables multi-granularity manipulation controlled by learnable modifiers (e.g., ``action class'', coarse- and fine-grained human body parts including ``person, head, body, arms, legs'', and kinematic information ``speed''), boosting motion reconstruction for single human and multi-human scenarios.}
    \label{top}
\end{center}
}]

\footnotetext[1]{Corresponding author.}

\maketitle
\input{sec/0_abstract}    
\input{sec/1_intro}

\input{sec/2_relatedwork}

\input{sec/3_approach}

\input{sec/4_experiments}

\input{sec/5_conclusion}

{
    \small
    \bibliographystyle{ieeenat_fullname}
    \bibliography{main}
}

\end{document}

%% file: preamble.tex
\usepackage[dvipsnames]{xcolor}

\usepackage{url}            
\usepackage{booktabs}       
\usepackage{amsfonts}       
\usepackage{nicefrac}       
\usepackage{microtype}      
\usepackage{graphicx}
\usepackage{amsmath}
\usepackage{bm}
\usepackage{adjustbox}
\usepackage{subcaption}
\usepackage{siunitx}
\usepackage{colortbl}
\usepackage{color}
\usepackage{pifont}
\usepackage{array}           
\usepackage{tabu}            
\usepackage{multirow}
\usepackage{makecell}
\usepackage{amssymb}
\definecolor{ggreen}{HTML}{58a55c}

\definecolor{Gray}{gray}{0.95}
\definecolor{Gray7}{gray}{0.75}

\definecolor{bestcolor}{gray}{.9}

\definecolor{mypink}{rgb}{.99,.95,.97}  
\definecolor{myblue}{rgb}{.93,.97,.99}

%% file: sec/0_abstract.tex
\begin{abstract}

The 3D Human Pose Estimation (3D HPE) task uses 2D images or videos to predict human joint coordinates in 3D space. Despite recent advancements in deep learning-based methods, they mostly ignore the capability of coupling accessible texts and naturally feasible knowledge of humans, missing out on valuable implicit supervision to guide the 3D HPE task. Moreover, previous efforts often study this task from the perspective of the whole human body, neglecting fine-grained guidance hidden in different body parts. To this end, we present a new Fine-Grained Prompt-Driven Denoiser based on a diffusion model for 3D HPE, named \textbf{FinePOSE}.
It consists of three core blocks enhancing the reverse process of the diffusion model: (1) Fine-grained Part-aware Prompt learning (FPP) block constructs fine-grained part-aware prompts via coupling accessible texts and naturally feasible knowledge of body parts with learnable prompts to model implicit guidance. (2) Fine-grained Prompt-pose Communication (FPC) block establishes fine-grained communications between learned part-aware prompts and poses to improve the denoising quality. (3) Prompt-driven Timestamp Stylization (PTS) block integrates learned prompt embedding and temporal information related to the noise level to enable adaptive adjustment at each denoising step. Extensive experiments on public single-human pose estimation datasets show that FinePOSE outperforms state-of-the-art methods. We further extend FinePOSE to multi-human pose estimation. Achieving 34.3mm average MPJPE on the EgoHumans dataset demonstrates the potential of FinePOSE to deal with complex multi-human scenarios.
Code is available at \url{https://github.com/PKU-ICST-MIPL/FinePOSE_CVPR2024}.

\end{abstract}
\vspace{-12pt}

%% file: sec/1_intro.tex
\section{Introduction}
\label{sec:intro}
Given monocular 2D images or videos, 3D Human Pose Estimation (3D HPE) aims to predict the positions of human body joints in 3D space. It is vital in various applications, including self-driving \cite{zanfir2023hum3dil,zheng2022multi}, sports analysis \cite{ingwersen2023sportspose,xu2022finediving,rematas2018soccer}, abnormal detection~\cite{epstein2020oops,xu2022unintentional}, and human-computer interaction ~\cite{ng2020you2me,wang2021synthesizing,hassan2021populating}.
Considering the expensive computational costs of directly obtaining 3D human poses from 2D contents, 3D HPE is usually decomposed into two stages: 1) detecting 2D keypoints in images or videos~\cite{cao2017realtime-cvpr,chen2018CPN-cvpr,sun2019deep,newell2016stacked}, and 2) mapping 2D keypoints to 3D human poses~\cite{d3dp-shan2023diffusion-iccv,gong2023diffpose-cvpr,chen2021anatomy,yu2023gla,zhang2022mixste}. In this work, we mainly focus on the second stage, estimating 3D human poses given 2D keypoints.

Existing monocular 3D HPE methods~\cite{wu2022c3p,zhang2022mixste,li2022mhformer,yu2023gla,zhu2023motionbert,d3dp-shan2023diffusion-iccv,zhao2023poseformerv2,gong2023diffpose-cvpr,xu2021graph-cvpr,nie2023lifting-ijcv,li2019mdn-cvpr,sharma2019cvae-cvpr,cai2019exploiting-iccv,li2020gan-bmvc,wehrbein2021NF-iccv,zou2021modulated-iccv,chen2021anatomy,oikarinen2021graphmdn-ijcnn} usually have three challenges as follows: 1) Uncertainty: the depth ambiguity inherently exists in the mapping from 2D skeletons to 3D ones (one-to-many);
2) Complexity: flexible human body structure, complex inter-joint relationships, and a high limb freedom degree lead to self-occlusion or rare and complicated poses;
3) Generalizability: current publicly available 3D HPE datasets have limited action classes, and thus, the models trained on such data are prone to overfitting and difficult to generalize to more diverse action classes.

To address these issues, we consider improving the 3D HPE model performance by enhancing the input information. We found that existing methods ignore accessible texts and naturally feasible knowledge of humans while they promise to provide the model with more guidance. We explicitly utilize (1) the action class of human poses, (2) kinematic information ``speed'', and (3) the way that different human body parts (e.g., person, head, body, arms, and legs) move in human activities to build \textbf{\textit{fine-grained part-aware prompts}} for the reconstruction task.
Specifically, we incorporate a fine-grained part-aware prompt learning mechanism into our framework to drive 3D human pose estimation via vision-language pre-trained models. It is well known that text prompts play a crucial role in various downstream tasks for vision-language pre-training models (e.g., CLIP~\cite{radford2021learning-icml}). However, manually designing prompt templates is expensive and cannot ensure that the final prompt is optimal for the 3D HPE task. Thus, we create a new fine-grained part-aware prompt learning mechanism that adaptively learns modifiers for different human body parts to precisely describe their movements from multiple granularities, including action class, speed, the whole person, and fine-grained human body parts. This new mechanism, coupled with diffusion models, possesses controllable high-quality generation capability, which is beneficial in addressing the challenges of the 3D human pose estimation task.

In this work, we propose a Fine-grained Prompt-driven Denoiser (\textbf{\textit{FinePOSE}}) based on diffusion models for 3D human pose estimation, in \cref{top}, which is composed of a fine-grained part-aware prompt learning (\textbf{\textit{FPP}}) block, fine-grained prompt-pose communication (\textbf{\textit{FPC}}) block, and prompt-driven timestamp stylization (\textbf{\textit{PTS}}) block. Concretely, the FPP block encodes three kinds of information about the human pose, including action class, coarse- and fine-grained parts of humans like ``person, head, body, arms, legs'', and kinematic information ``speed'', and integrates them with pose features for serving subsequent processes. Then, the FPC block injects fine-grained part-aware prompt embedding into noise 3D poses to establish fine-grained communications between learnable part-aware prompts and poses for enhancing the denoising capability. To handle 3D poses with different noise levels, the PTS block introduces the timestamp coupled with fine-grained part-aware prompt embedding into the denoising process to enhance its adaptability and refine the prediction at each noise level.

Our contributions can be summarized as follows:
\begin{itemize}
    \item We propose a new fine-grained part-aware prompt learning mechanism coupled with diffusion models that possesses human body part controllable high-quality generation capability, beneficial to the 3D human pose estimation task.
    \item Our FinePOSE encodes multi-granularity information about action class, coarse- and fine-grained human parts, and kinematic information, and establishes fine-grained communications between learnable part-aware prompts and poses for enhancing the denoising capability.
    \item Extensive experiments illustrate that our FinePOSE obtains substantial improvements on Human3.6M and MPI-INF-3DHP datasets and achieves state-of-the-art. More experiments on EgoHumans demonstrate the potential of FinePOSE to deal with complex multi-human scenarios.
\end{itemize}

%% file: sec/2_relatedwork.tex
\section{Related Work}
\label{sec:formatting}

\textbf{Diffusion Models.} Diffusion models~\cite{ho2020denoising-nips,song2020denoising,nichol2021improved,song2020score} are a kind of generative models that sequentially add a series of noise with different levels to the raw data, gradually transforming it from an original data distribution to a noisy distribution, and subsequently reconstructing the original data by denoising.
Diffusion models have strong capabilities in many applications, from 2D image or video generation/editing~\cite{austin2021structured, avrahami2023blended, bao2022analytic, kim2023diffusion, yu2023video} to 3D human pose estimation/generation~\cite{zhang2022mixste,li2022mhformer, yu2023gla, zhu2023motionbert,d3dp-shan2023diffusion-iccv,zhao2023poseformerv2,gong2023diffpose-cvpr,xu2021graph-cvpr,nie2023lifting-ijcv,li2019mdn-cvpr}.
The 3D HPE task, for example, encounters various difficulties, including occlusions, limited training data, and inherent ambiguity in pose representations. Therefore, diffusion models' ability to generate high-fidelity 3D human poses makes them more suitable for 3D HPE.

\begin{figure*}[t]
  \centering
  \includegraphics[width=\linewidth]{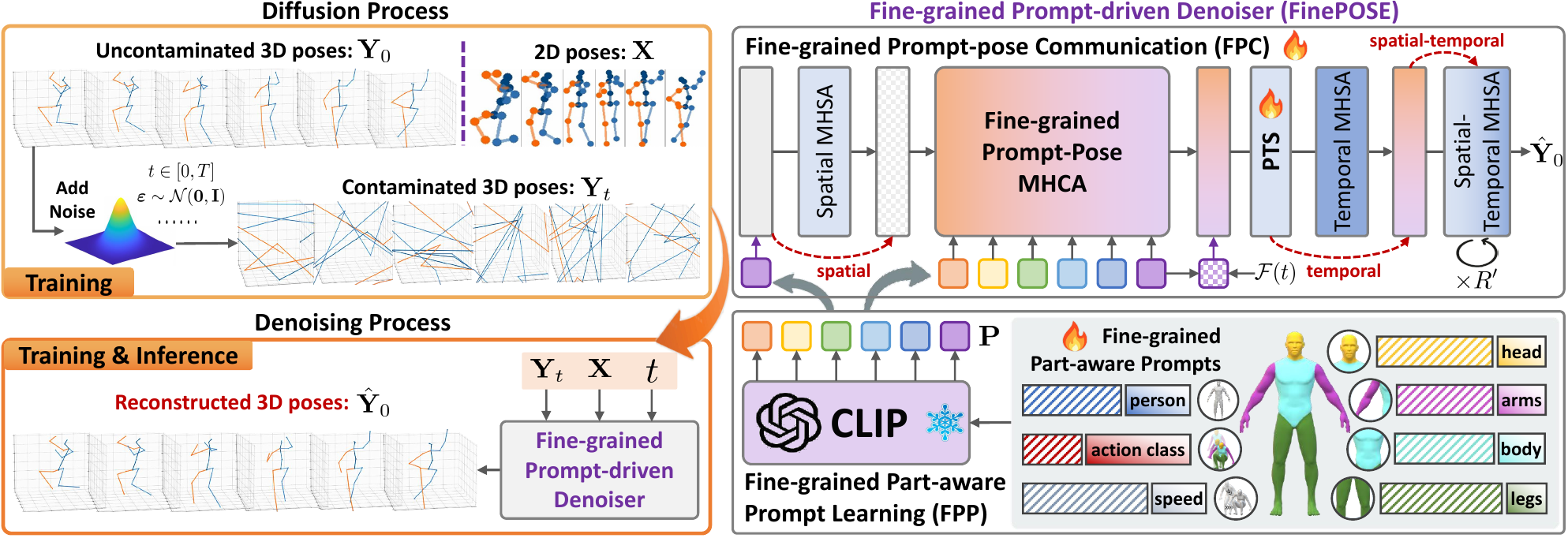}
    \vspace{-19pt}
  \caption{\textbf{The architecture of the proposed FinePOSE.}
  In the diffusion process, Gaussian noise is gradually added to the ground-truth 3D poses $\mathbf{Y}_0$, generating the noisy 3D poses $\mathbf{Y}_t$ for the timestamp $t$. In the denoising process, $\mathbf{Y}_t$, $\mathbf{X}$ and $t$ are fed to fine-grained prompt-driven denoiser $\mathcal{D}$ to reconstruct pure 3D poses $\hat{\mathbf{Y}}_0$. $\mathcal{D}$ is composed of a Fine-grained Part-aware Prompt learning (FPP) block, a Fine-grained Prompt-pose Communication (FPC) block, and a Prompt-driven Timestamp Stylization (PTS) block, where FPP provides more precise guidance for all human part movements, FPC establishes fine-grained communications between learnable prompts and poses for enhancing the denoising capability, and PTS integrates learned prompt embedding and current timestamp for refining the prediction at each noise level.
  }
  \label{fig:arch}
  \vspace{-10pt}
\end{figure*}

\noindent\textbf{3D Human Pose Estimation.} Considering that extracting 2D human skeletons from videos or images requires expensive costs, the 3D human pose estimation task is usually divided into two phases: (1) estimating 2D positions of human joints from images or videos~\cite{chen2018CPN-cvpr,cao2017realtime-cvpr, wang2020deep, mao2022poseur}, and (2) mapping 2D positions to the 3D space to estimate the 3D positions of human joints ~\cite{zhang2022mixste,li2022mhformer, yu2023gla, zhu2023motionbert,d3dp-shan2023diffusion-iccv,zhao2023poseformerv2,gong2023diffpose-cvpr,xu2021graph-cvpr,nie2023lifting-ijcv,li2019mdn-cvpr,sharma2019cvae-cvpr,cai2019exploiting-iccv,li2020gan-bmvc,wehrbein2021NF-iccv,zou2021modulated-iccv,chen2021anatomy,oikarinen2021graphmdn-ijcnn}. In this work, we focus on the second phase.
Early, TCN~\cite{pavllo20193d} used a fully convolutional network based on dilated temporal convolutions over 2D keypoints to estimate 3D poses in video. SRNet~\cite{zeng2020srnet} proposed a split-and-recombine approach, leading to appreciable improvements in predicting rare and unseen poses. Anatomy~\cite{chen2021anatomy} decomposed the task into bone direction prediction and bone length prediction, from which the 3D joint locations can be derived entirely. Recently, MixSTE~\cite{zhang2022mixste} used temporal and spatial transformers alternately to obtain better spatio-temporal features. MotionBERT~\cite{zhu2023motionbert} proposed a pretraining stage to recover the underlying 3D motion from noisy partial 2D observations. GLA-GCN~\cite{yu2023gla} globally modeled the spatio-temporal structure for 3D human pose estimation. D3DP~\cite{d3dp-shan2023diffusion-iccv} proposed the joint-level aggregation strategy to benefit from all generated poses. Unlike previous methods, our approach proposes a new fine-grained part-aware prompt learning mechanism coupled with diffusion models that possess controllable, high-quality generation capability of human body parts, which benefits the 3D human pose estimation task.

\noindent\textbf{Prompt Learning.} Prompt learning has been widely used in the computer vision community~\cite{zhou2022learning-ijcv,zhou2022conditional-cvpr,lu2022prompt-cvpr,derakhshani2022variational-arxiv}. Typically, CoOp~\cite{zhou2022learning-ijcv} utilized a continuous prompt optimization from downstream data instead of hand-craft design, the pioneering work that brings prompt learning to adapt pre-trained vision language models. CoCoOp~\cite{zhou2022conditional-cvpr} extended CoOp by learning image conditional prompts to improve generalization. ProDA~\cite{lu2022prompt-cvpr} learned a prompt distribution over the output embedding space. VPT~\cite{derakhshani2022variational-arxiv} introduced variational prompt tuning by combining a base learned prompt with a residual vector sampled from an instance-specific underlying distribution. PointCLIPV2~\cite{zhu2023pointclip} combined CLIP~\cite{radford2021learning-icml} with GPT~\cite{liu2023gpt} to be a unified 3D open-world learner. Unlike the above methods, we propose a new fine-grained part-aware prompt learning mechanism, which encodes multi-granularity information about action class, coarse- and fine-grained human parts, and kinematic data, and establishes fine-grained communications between learnable part-aware prompts and poses for enhancing the denoising capability.

%% file: sec/3_approach.tex
\section{The Proposed Approach: FinePOSE} \label{sec:approach}

Given a 2D keypoints sequence $\mathbf{X}\in\mathbb{R}^{N\times J\times2}$, constructed by $N$ frames with $J$ joints in each, the proposed approach is formulated to predict the 3D pose sequence $\mathbf{Y}\in\mathbb{R}^{N\times J\times3}$.
Considering the high-quality generation capability of the text-controllable denoising process of diffusion models, we develop a Fine-grained Prompt-driven Denoiser (FinePOSE) $\mathcal{D}$ for 3D human pose estimation. FinePOSE generates accurate 3D human poses enhanced by three core blocks: Fine-grained Part-aware Prompt learning (FPP), Fine-grained Prompt-pose Communication (FPC), and Prompt-driven Timestamp Stylization (PTS) blocks.

\subsection{Diffusion-Based 3D Human Pose Estimation}
Diffusion models are generative models that model the data distribution in the form of $p_{\theta}(\mathbf{Y}_{0})\!:=\!\int\!p_{\theta}(\mathbf{Y}_{0:T})d\mathbf{Y}_{1:T}$ through chained diffusion and reverse (denoising) processes.
The diffusion process gradually adds Gaussian noise into the ground truth 3D pose sequence $\mathbf{Y}_{0}$ to corrupt it into an approximately Gaussian noise $\mathbf{Y}_{t} (t\!\to\!T)$ using a variance schedule $\{\beta_t\}_{t=1}^T$, which can be formulated as
\begin{equation}\label{eq1}
q\left(\mathbf{Y}_{t}\mid\mathbf{Y}_{0}\right):=\sqrt{\bar{\alpha}_{t}} \mathbf{Y}_{0}+\epsilon\sqrt{1-\bar{\alpha}_{t}},
\end{equation}
where $\bar{\alpha}_{t}\!:=\!{\textstyle\prod_{s=0}^t}\alpha_s$ and $\alpha_{t}\!:=\!1\!-\!\beta_t$.
Afterward, the denoising process reconstructs the uncontaminated 3D poses by a denoiser $\mathcal{D}$. Since the degraded data is well approximated by a Gaussian distribution after the diffusion process, we can obtain initial 3D poses $\mathbf{Y}_T$ by sampling noise from a unit Gaussian.
Passing $\mathbf{Y}_T (t\!=\!T)$ to the denoiser $\mathcal{D}$, we obtain $\hat{\mathbf{Y}}_0$ that is thereafter used to generate the noisy 3D poses $\hat{\mathbf{Y}}_{t\!-\!1}$ as inputs to the denoiser $\mathcal{D}$ at timestamp $t\!-\!1$ via DDIM~\cite{song2020denoising},  which can be formulated as
\begin{equation} \label{eq:DDIM}
    \mathbf{Y}_{t\!-\!1}=\sqrt{\bar{\alpha}_{t\!-\!1}}\hat{\mathbf{Y}}_0\!+\!\epsilon_t\sqrt{1\!-\!\bar{\alpha}_{t\!-\!1}\!-\!\sigma^2_t}\!+\!\sigma_t\epsilon,
\end{equation}
where $t$ is from $T$ to $1$, $\epsilon\!\sim\!\mathcal{N}(0,\mathbf{I})$ is standard Gaussian noise independent of $\mathbf{Y}_t$, and
\begin{subequations}
\begin{align}
\epsilon_t&=\left(\mathbf{Y}_t\!-\!\sqrt{\bar{\alpha}_t}\cdot \hat{\mathbf{Y}}_0\right)/\sqrt{1\!-\!\bar{\alpha}_t}, \\
\sigma_t&=\sqrt{\left(1\!-\!\bar{\alpha}_{t\!-\!1}\right)/\left(1\!-\!\bar{\alpha}_t\right)}\cdot\sqrt{1\!-\!(\bar{\alpha}_t/\bar{\alpha}_{t\!-\!1})},
\end{align}
\end{subequations}
where $\epsilon_t$ is the noise at timestamp $t$, and $\sigma_t$ controls how stochastic the diffusion process is.

\subsection{Fine-grained Prompt-driven Denoiser}

\noindent\textbf{Fine-grained Part-aware Prompt Learning (FPP).}
To assist the reconstruction of pure 3D poses $\hat{\mathbf{Y}}_0$ from contaminated 3D poses $\mathbf{Y}_t$ with additional information, FinePOSE guides the denoising process with regular 2D keypoints $\mathbf{X}$, timestamp $t$, and fine-grained part-aware prompt embedding $\mathbf{P}$. We design the FPP block to learn $\mathbf{P}$.
It encodes three pose-related information in the prompt embedding space, including its action class, coarse- and fine-grained parts of humans like ``person, head, body, arms, legs'', and kinematic information ``speed''. Afterward, $\mathbf{P}$ is integrated with pose features for subsequent processes.

A learnable prompt embedding $\mathbf{P}\!=\!\{\bm{p}\}_{k=1}^K$ is with the shape of $K\times\!L\times\!D$, where $K$ denotes the number of text prompts, $L$ indicates the number of tokens in each text prompt, and $D$ is the dimension of token embedding.
Since the number of valid tokens is found to be three to four through the text encoder $\mathcal{E}_{\text{tx}}$, the first four tokens are taken as representations $\tilde{\bm{p}}_k$ for each text. Moreover, since modifiers help precisely describe the movements of human body parts, we design a learnable vector $\bm{r}_k\in\mathbb{R}^{(L_k-4)\times D}$ to wrap the representations as $\bm{p}_k$. The above can be formulated as
\begin{subequations}
\begin{align}
    \tilde{\bm{p}}_k&=\mathcal{E}_{\text{tx}}(\text{text}_k)[:4],\ k \in [1, K],\\
    \bm{p}_k&=\text{Concat}(\bm{r}_k, \tilde{\bm{p}}_k),
\end{align}
\end{subequations}
where $K\!=\!7$ and $\{\text{text}_k\}_{k=1}^7$ indicate $\{$person, [Action Class], speed, head, body, arms, legs$\}$. $\bm{r}_k$ is initialized with Gaussian distribution of $\mu\!\!=\!\!0$ and $\sigma\!\!=\!\!0.02$, and $\{L_k\}_{k=1}^7\!=\!\{7, 12, 10, 10, 10, 14, 14\}$, which sums to 77 regarding the text embedding dimension of CLIP~\cite{radford2021learning-icml}.
In short, the FPP block builds multi-granularity text prompts and learnable modifiers, providing precise guidance for each human body part, as shown in \cref{fig:arch}.

\noindent\textbf{Fine-grained Prompt-pose Communication (FPC).}
After obtaining fine-grained part-aware prompt embedding $\mathbf{P}$, we establish fine-grained communications between learned part-aware prompts and poses using the FPC block to improve the denoising quality. Specifically, when processing the noised 3D poses $\mathbf{Y}_t$, it injects prompt embedding $\mathbf{P}$, 2D keypoints $\mathbf{X}$, and timestamp $t$ within.

First, FPC integrates $\mathbf{Y}_t$ and guidance information (i.e., $\mathbf{X}$, $t$, and $\mathbf{P}$) by a series of concatenation and addition operations, as $\mathbf{Z}_t\!=\!\text{Concat}(\mathbf{Y}_t,\mathbf{X})\!+\!\mathbf{P}[L]\!+\!\mathcal{F}(t)$. $\mathcal{F}$ is the timestamp embedding network containing a sinusoidal function followed by two Linear layers connected by a GELU non-linearity. The timestep embedding adaptively adjusts the quantity of Gaussian noise additions. Since the denoiser $\mathcal{D}$ works iteratively, providing detailed information about the current timestamp $t$ is crucial for $\mathcal{D}$ to handle 3D poses containing different noise levels effectively. Then, $\mathbf{Z}_t$ is encoded by a spatial transformer, where the multi-head self-attention (MHSA) mechanism helps to focus on the fine-grained relationships between joints within each frame, obtaining $\mathbf{Z}_t^s$.

To completely inject prompt embedding $\mathbf{P}$ into $\mathbf{Z}_t^s$, we implement a multi-head cross-attention model, where the \textit{query}, \textit{key}, and \textit{value} are as $\mathbf{Q}\!=\!\mathbf{W}_Q\mathbf{Z}_t^s$, $\mathbf{K}\!=\!\mathbf{W}_K\mathbf{P}$, $\mathbf{V}\!=\!\mathbf{W}_V\mathbf{P}$.
The \textit{value} is aggregated with cross-attention $\mathbf{A}$ to generate fine-grained prompt-driven pose features $\mathbf{Z}_t^{sp}$, achieving fine-grained prompt-pose communication. The mechanism can be formulated as
\begin{subequations}
\begin{align}
\mathbf{A}&=\text{softmax}(\mathbf{Q}\otimes\mathbf{K}^\top/\sqrt{d}),\\
\mathbf{Z}_t^{sp}&=\mathbf{A}\otimes\mathbf{V},\ \tilde{\mathbf{Z}}_t^{sp}=\mathcal{P}(\mathbf{Z}_t^{sp}),
\end{align}
\end{subequations}
where $d\!=\!D/H$ and $H$ is the number of attention heads. $\mathcal{P}$ indicates the PTS block that bring timestamp $t$ into the generation process to obtain timestamp stylized output $\tilde{\mathbf{Z}}_t^{sp}$. On the other hand, to model inter-frame relationships between poses, $\tilde{\mathbf{Z}}_t^{sp}$ is encoded using a temporal transformer via MHSA to obtain  $\tilde{\mathbf{Z}}_t^{spf}$.
Finally, we utilize a spatial-temporal transformer accompanied by permutation operations between spatial and temporal dimensions to extract more compact fine-grained prompt-driven pose features from $\tilde{\mathbf{Z}}_t^{spf}$, which are decoded as the predicted 3D poses $\hat{\mathbf{Y}}_0$.

\noindent\textbf{Prompt-driven timestamp Stylization (PTS).} As mentioned, providing timestamp embedding to the denoising process is critical for handling 3D poses with different noise levels. Therefore, inspired by Motiondiffuse~\cite{zhang2022motiondiffuse-arXiv}, we introduce the PTS block that explicitly embeds timestamp $t$ by positional embedding~\cite{vaswani2017attention-nips} and sums it with the learnable prompt embedding $\mathbf{P}$ obtained by the FPP block, as $\bm{v}\!=\!\mathbf{P}[L]\!+\!\mathcal{F}(t)$. Given the intermediate output $\mathbf{Z}_t^{sp}$ of the FPC block, the PTS block calculates  $\tilde{\mathbf{Z}}_t^{sp}\!=\!\mathbf{Z}_t^{sp}\cdot\psi_w(\phi(\bm{v}))\!+\!\psi_b(\phi(\bm{v}))$, where $\psi_b, \psi_w, \phi$ are three different linear projections, and $(\cdot)$ is the Hadamard product.

\begin{table*}[t]
\adjustbox{width=\linewidth}
  {
  \setlength{\tabcolsep}{11.5pt}
  
    \begin{tabular}{lcccccccl}
    \toprule
    \multirow{2}{*}{Method} 
    & \multirow{2}{*}{$N$}  
    & \multicolumn{3}{c}{Human3.6M (DET)} 
    & \multicolumn{3}{c}{Human3.6M (GT)}
    & \multirow{2}{*}{Year} \\ 
    \cmidrule(lr){3-5} \cmidrule(lr){6-8}
    &
    & Detector & MPJPE~$\textcolor{blue}{\downarrow}$ & P-MPJPE~$\textcolor{blue}{\downarrow}$
    & Detector & MPJPE~$\textcolor{blue}{\downarrow}$ & P-MPJPE~$\textcolor{blue}{\downarrow}$\\ 
    \midrule
    TCN~\cite{pavllo20193d}  & 243 & CPN & 46.8 & 36.5 & GT & 37.8 &  / & \textcolor{gray}{CVPR'19} \\
    Anatomy~\cite{chen2021anatomy} & 243 & CPN & 44.1 & 35.0 & GT & 32.3 & /  & \textcolor{gray}{CSVT'21} \\
    P-STMO~\cite{shan2022p} & 243 & CPN & 42.8 & 34.4 & GT & 29.3 &  / & \textcolor{gray}{ECCV'22} \\
    MixSTE~\cite{zhang2022mixste} & 243 & HRNet & 39.8 & 30.6 & GT & 21.6 &  / & \textcolor{gray}{CVPR'22}\\
    PoseFormerV2~\cite{zhao2023poseformerv2} & 243 & CPN & 45.2 & 35.6 & GT & 35.5 & /  & \textcolor{gray}{CVPR'23} \\
    MHFormer~\cite{li2022mhformer} & 351 & CPN & 43.0 & 34.4 & GT & 30.5  & /  & \textcolor{gray}{CVPR'22}\\
    Diffpose~\cite{gong2023diffpose-cvpr} & 243 & CPN & 36.9 & \underline{28.7} & GT & 18.9 &  / & \textcolor{gray}{CVPR'23} \\
    GLA-GCN~\cite{yu2023gla} & 243 & CPN & 44.4 & 34.8 & GT & 21.0 & 17.6 & \textcolor{gray}{ICCV'23}\\
    ActionPrompt~\cite{zheng2023actionprompt-icme} & 243 & CPN & 41.8 & 29.5 & GT & 22.7 & /  & \textcolor{gray}{ICME'23}\\
    MotionBERT~\cite{zhu2023motionbert} & 243 & SH & 37.5 & / & GT & \underline{16.9} &  / & \textcolor{gray}{ICCV'23} \\
    D3DP~\cite{Shan_2023_ICCV} & 243 & CPN & \underline{35.4} & \underline{28.7} & GT & 18.4 & / & \textcolor{gray}{ICCV'23} \\
    \midrule
    \rowcolor{Gray}\textbf{FinePOSE} (Ours) & 243 & CPN &  \textbf{31.9} & \textbf{25.0} & GT & \textbf{16.7} & \textbf{12.7} & \\
    \rowcolor{Gray} & & & \textcolor{blue}{(-3.5)} &  \textcolor{blue}{(-3.7)} & & \textcolor{blue}{(-0.2)} & \textcolor{blue}{(-4.9)} & \\
    \bottomrule
    \end{tabular}
}
\vspace{-7pt}
\caption{\textbf{Quantitative comparison with the state-of-the-art 3D human pose estimation methods on the Human3.6M dataset.} $N$: the number of input frames. CPN, HRNet, SH: using CPN~\cite{chen2018CPN-cvpr}, HRNet~\cite{sun2019deep}, and SH~\cite{newell2016stacked} as the 2D keypoint detectors to generate the inputs. GT: using the ground truth 2D keypoints as inputs. The best and second-best results are highlighted in \textbf{bold} and \underline{underlined} formats.}
\vspace{-10pt}
\label{tab:SOTA-H3.6M}
\end{table*}

\subsection{Training \& Inference}
\textbf{Training.} The contaminated 3D poses $\mathbf{Y}_t$ is sent to a fine-grained prompt-driven denoiser $\mathcal{D}$ to reconstruct the 3D poses $\hat{\mathbf{Y}}_0\!=\!\mathcal{D}(\mathbf{Y}_t, \mathbf{X}, t, \mathbf{P})$ without noise. The entire framework is optimized by minimizing the MSE loss $\|\mathbf{Y}_0-\hat{\mathbf{Y}}_0\|_2$.

\noindent\textbf{Inference.} Since the distribution of $\mathbf{Y}_T$ is nearly an isotropic Gaussian distribution, we sample $H$ initial 3D poses $\{\mathbf{Y}_T^h\}_{h=1}^H$ from a unit Gaussian. After passing them to the denoiser $\mathcal{D}$, we obtain $H$ feasible 3D pose hypotheses $\{\hat{\mathbf{Y}}_0^h\}_{h=1}^H$.
Each hypothesis $\hat{\mathbf{Y}}_0^h$ is used to generate the noisy 3D poses $\hat{\mathbf{Y}}_{t-1}^h$ as inputs to the denoiser $\mathcal{D}$ for the next timestamp $t\!-\!1$. Then, we regenerate $\{\hat{\mathbf{Y}}_0^h\}_{h=1}^H$ using $\{\hat{\mathbf{Y}}_{t\!-\!1}^h\}_{h=1}^H$ as inputs to the denoiser $\mathcal{D}$ for the next timestamp $t\!-\!2$. Analogously, this process iterates $M$ times starting from the timestamp $T$, so each iteration $m \!\in\![1, M]$ is with the timestamp $t\!=\!T(1\!-\!\frac{m}{M})$.
Following Joint-Wise Reprojection-Based Multi-Hypothesis Aggregation (JPMA) in~\cite{d3dp-shan2023diffusion-iccv}, we reproject $\{\hat{\mathbf{Y}}_0^h\}_{h=1}^H$ to the 2D camera plane using known or estimated intrinsic camera parameters and then choose joints with minimum projection errors with the input $\mathbf{X}$, as
\begin{subequations}
    \begin{align}
h'&=\mathop{\arg\min}\limits_{h\in[1,H]}\|\mathcal{P}_R(\hat{\mathbf{Y}}_0^h)[j]-\mathbf{X}[j]\|_2,\\
    \hat{\mathbf{Y}}_0[j]&=\hat{\mathbf{Y}}_0^{h'}[j],\ j\in[1,J],
    \end{align}
\end{subequations}
where $\mathcal{P}_R$ is the reprojection function, $j$ is the index of joints, and $h'$ indicates the index of selected hypothesis.
JPMA enables us to select joints from distinct hypotheses automatically to form the final prediction $\hat{\mathbf{Y}}_0$.

\subsection{Extension to 3D Multi-Human Pose Estimation}
\label{subsec:multipp}
We append a post-integration to FinePOSE to apply for the multi-human scenario, avoiding incorporating extra computational cost. Specifically, given a multi-human 2D keypoints sequence $\mathbf{X}_{\text{mul}}\in \mathbb{R}^{C\times N\times J\times2}$, which involves $C$ human characters, FinePOSE first predicts $\hat{\mathbf{Y}}_0^{\tiny c}$ for each character $c \in [1, C]$.
Considering that some characters may temporarily leave the camera field of view, their positions in those frames are set as zeros to ensure synchronization of all characters' states in $\mathbf{X}_{\text{mul}}$.
Next, we integrate $\{\hat{\mathbf{Y}}_0^{\tiny c}\}^C_{c=1}$ by stacking over the character dimension, obtaining the final prediction $\hat{\mathbf{Y}}_0^C \in \mathbb{R}^{C\times N\times J\times3}$.

%% file: sec/4_experiments.tex
\section{Experiments}

\begin{table*}[t]
\adjustbox{width=\linewidth}
  {
  \setlength{\tabcolsep}{4.3pt}
    \begin{tabular}{lcccccccccccccccl}
    \toprule
    \multirow{2}{*}{Method / MPJPE~$\textcolor{blue}{\downarrow}$} 
    & \multicolumn{16}{c}{Human3.6M (DET)} \\ 
    \cmidrule(lr){2-17}
    & Dir. & Disc. & Eat & Greet & Phone & Photo & Pose & Pur. & Sit & SitD. & Smoke & Wait & WalkD. & Walk & WalkT. & Avg \\ 
    \midrule
    TCN~\cite{pavllo20193d}  & 45.2 & 46.7 & 43.3 & 45.6 & 48.1 & 55.1 & 44.6 & 44.3 & 57.3 & 65.8 & 47.1 & 44.0 & 49.0 & 32.8 & 33.9 & 46.8\\
    SRNet~\cite{zeng2020srnet} & 46.6 & 47.1 & 43.9 & 41.6 & 45.8 & 49.6 & 46.5 & 40.0 & 53.4 & 61.1 & 46.1 & 42.6 & 43.1 & 31.5 & 32.6 & 44.8\\
    RIE~\cite{shan2021improving} & 40.8 & 44.5 & 41.4 & 42.7 & 46.3 & 55.6 & 41.8 & 41.9 & 53.7 & 60.8 & 45.0 & 41.5 & 44.8 & 30.8 & 31.9 & 44.3\\
    Anatomy~\cite{chen2021anatomy} & 41.4 & 43.5 & 40.1 & 42.9 & 46.6 & 51.9 & 41.7 & 42.3 & 53.9 & 60.2 & 45.4 & 41.7 & 46.0 & 31.5 & 32.7 & 44.1\\
    P-STMO~\cite{shan2022p} & 38.9 & 42.7 & 40.4 & 41.1 & 45.6 & 49.7 & 40.9 & 39.9 & 55.5 & 59.4 & 44.9 & 42.2 & 42.7 & 29.4 & 29.4 & 42.8 \\
    MixSTE~\cite{zhang2022mixste} & 36.7 & 39.0 & 36.5 & 39.4 & 40.2 & 44.9 & 39.8 & 36.9 & 47.9 & 54.8 & 39.6 & 37.8 & 39.3 & 29.7 & 30.6 & 39.8\\
    PoseFormerV2~\cite{zhao2023poseformerv2} & - & - & - & - & - & - & - & - & - & - & - & - & - & - & - & 45.2\\
    MHFormer~\cite{li2022mhformer} & 39.2 & 43.1 & 40.1 & 40.9 & 44.9 & 51.2 & 40.6 & 41.3 & 53.5 & 60.3 & 43.7 & 41.1 & 43.8 & 29.8 & 30.6 & 43.0\\
    Diffpose~\cite{gong2023diffpose-cvpr} & 33.2 & 36.6 & 33.0 & 35.6 & 37.6 & 45.1 & 35.7 & 35.5 & 46.4 & 49.9 & 37.3 & 35.6 & 36.5 & 24.4 & \underline{24.1} & 36.9\\
    GLA-GCN~\cite{yu2023gla} & 41.3 & 44.3 & 40.8 & 41.8 & 45.9 & 54.1 & 42.1 & 41.5 & 57.8 & 62.9 & 45.0 & 42.8 & 45.9 & 29.4 & 29.9 & 44.4\\
    ActionPrompt~\cite{zheng2023actionprompt-icme} & 37.7 & 40.2 & 39.8 & 40.6 & 43.1 & 48.0 & 38.8 & 38.9 & 50.8 & 63.2 & 42.0 & 40.0 & 42.0 & 30.5 & 31.6 & 41.8\\
    MotionBERT~\cite{zhu2023motionbert} & 36.1 & 37.5 & 35.8 & \underline{32.1} & 40.3 & 46.3 & 36.1 & 35.3 & 46.9 & 53.9 & 39.5 & 36.3 & 35.8 & 25.1 & 25.3 & 37.5 \\
    D3DP~\cite{Shan_2023_ICCV} & \underline{33.0} & \underline{34.8} & \underline{31.7} & 33.1 & \underline{37.5} & \underline{43.7} & \underline{34.8} & \underline{33.6} & \underline{45.7} & \underline{47.8} & \underline{37.0} & \underline{35.0} & \underline{35.0} &  \underline{24.3} &  \underline{24.1} &  \underline{35.4}\\
    \midrule
    \rowcolor{Gray}\textbf{FinePOSE}~(Ours)& \textbf{31.4} & \textbf{31.5} & \textbf{28.8} & \textbf{29.7} & \textbf{34.3} & \textbf{36.5} & \textbf{29.2} & \textbf{30.0} & \textbf{42.0} & \textbf{42.5} & \textbf{33.3} & \textbf{31.9} & \textbf{31.4} & \textbf{22.6} & \textbf{22.7} & \textbf{31.9}\\
    
     \rowcolor{Gray} & \textcolor{blue}{(-1.6)} & \textcolor{blue}{(-3.3)} & \textcolor{blue}{(-2.9)} & \textcolor{blue}{(-2.4)} & \textcolor{blue}{(-3.2)} & \textcolor{blue}{(-7.2)} & \textcolor{blue}{(-5.6)} & \textcolor{blue}{(-3.6)} & \textcolor{blue}{(-3.7)} & \textcolor{blue}{(-5.3)} & \textcolor{blue}{(-3.7)} & \textcolor{blue}{(-3.1)} & \textcolor{blue}{(-3.6)} & \textcolor{blue}{(-1.7)} & \textcolor{blue}{(-1.4)} & \textcolor{blue}{(-3.5)}\\
    \bottomrule
    \end{tabular}
}
\vspace{-7pt}
\caption{\textbf{Quantitative comparison with the state-of-the-art 3D human pose estimation methods on the Human3.6M dataset using 2D keypoint detectors to generate the inputs.} \textit{Dir.}, \textit{Disc.},$\cdots$, and \textit{WalkT.} correspond to 15 action classes. \textit{Avg} indicates the average MPJPE among 15 action classes. The best and second-best results are highlighted in \textbf{bold} and \underline{underlined} formats.}
\vspace{-5pt}
\label{tab:quantitive-H3.6M-det}
\end{table*}

\begin{table}[t]
\adjustbox{width=\linewidth}
  {
  \setlength{\tabcolsep}{5pt}
  
    \begin{tabular}{llcccl}
    \toprule
    \multirow{2}{*}{Method} 
    & \multirow{2}{*}{$N$} 
    & \multicolumn{3}{c}{MPI-INF-3DHP} 
    & \multirow{2}{*}{Year} \\ 
    \cmidrule(lr){3-5}
    & 
    & PCK$\textcolor{red}{\uparrow}$ & AUC$\textcolor{red}{\uparrow}$ & MPJPE~$\textcolor{blue}{\downarrow}$ \\ 
    \midrule
    TCN~\cite{pavllo20193d}  & 81 & 86.0 & 51.9 & 84.0 & \textcolor{gray}{CVPR'19} \\
    Anatomy~\cite{chen2021anatomy} & 81 & 87.9 & 54.0 & 78.8 & \textcolor{gray}{CSVT'21} \\
    P-STMO~\cite{shan2022p} & 81 & 97.9 & 75.8 & 32.2 & \textcolor{gray}{ECCV'22} \\
    MixSTE~\cite{zhang2022mixste} & 27 & 94.4 & 66.5 & 54.9 & \textcolor{gray}{CVPR'22}\\
    PoseFormerV2~\cite{zhao2023poseformerv2} & 81 & 97.9 & 78.8 & \underline{27.8} & \textcolor{gray}{CVPR'23} \\
    MHFormer~\cite{li2022mhformer} & 9 & 93.8 & 63.3 & 58.0 & \textcolor{gray}{CVPR'22}\\
    Diffpose~\cite{gong2023diffpose-cvpr} & 81 & 98.0 & 75.9 & 29.1 & \textcolor{gray}{CVPR'23} \\
    GLA-GCN~\cite{yu2023gla} & 81 & \underline{98.5} & \underline{79.1} & \underline{27.8} & \textcolor{gray}{ICCV'23}\\
    D3DP~\cite{Shan_2023_ICCV} & 243 & 98.0 & \underline{79.1} & 28.1 & \textcolor{gray}{ICCV'23} \\
    \midrule
    \rowcolor{Gray}\textbf{FinePOSE} (Ours) & 243 & \textbf{98.9} & \textbf{80.0} & \textbf{26.2} & \\
    \rowcolor{Gray} &  & \textcolor{red}{(+0.4)} & \textcolor{red}{(+0.9)} & \textcolor{blue}{(-1.6)} & \\
    
    \bottomrule
    \end{tabular}
}
\vspace{-7pt}
\caption{\small \textbf{Quantitative comparison with the state-of-the-art 3D human pose estimation methods on the MPI-INF-3DHP dataset using ground truth 2D keypoints as inputs.} $N$: the number of input frames. The best and second-best results are highlighted in \textbf{bold} and \underline{underlined} formats.}
\vspace{-12pt}
\label{tab:SOTA-3DHP}
\end{table}

\subsection{Datasets and Metrics}
\textbf{Human3.6M}~\cite{ionescu2013human3} is a widely used benchmark dataset in human pose estimation tasks, which provides a large-scale collection of accurate 3D joint annotations on diverse human activities. Human3.6M consists of 3.6 million RGB images, captured from multiple camera views, of 11 professional actors performing 15 activities, e.g., walking, running, and jumping. Following previous efforts~\cite{pavllo20193d,li2022mhformer,Shan_2023_ICCV}, our FinePOSE is trained on five subjects (S1, S5, S6, S7, S8) and evaluated on two subjects (S9, S11). We calculate the mean per joint position error (i.e., MPJPE) to measure the average Euclidean distance in millimeters between the ground truth and estimated 3D joint positions for evaluation. We also report procrustes MPJPE (i.e., P-MPJPE) that calculates MPJPE after aligning the estimated poses to the ground truth using a rigid transformation.

\noindent\textbf{MPI-INF-3DHP}~\cite{mehta2017monocular} provides synchronized RGB video sequences with accurate 3D joint annotations for 3D human pose estimation. It comprises 8 activities conducted by 8 actors in the training set, while the test set encompasses 7 activities. We calculate MPJPE, the percentage of correctly estimated keypoints (i.e., PCK) within a 150mm range, and the area under the curve (i.e., AUC).

\noindent\textbf{EgoHumans}~\cite{Khirodkar_2023_ICCV} collects multi-human ego-exo videos covering 7 sports activities. Recently, a subset of 2D to 3D keypoints annotations has been released covering tagging, lego-assembling, and fencing. It contains 105 RGB videos taken by ego cameras. Between 1 and 3 human characters appear in each video, resulting in a total of 238 subsequences. We report the average MPJPE per video.

\subsection{Implementation Details}

We take MixSTE~\cite{zhang2022mixste} as the backbone of the denoiser $\mathcal{D}$ and CLIP as the frozen text encoder $\mathcal{E}_{\text{tx}}$. The numbers of MHSA-MLP-LN building blocks of the spatial, temporal, and spatio-temporal transformer in the FPC block are 1, 1, and 3. The training epoch in all the experiments below is 100, and the batch size is 4. We adopt AdamW optimizer with the momentum parameters of $\beta_1\!=\!$0.9, $\beta_2\!=\!$0.999, and the weight decay of 0.1. The learning rate starts from 6e$^{-5}$ and shrinks after each epoch with a factor of 0.993. For fair comparisons, we set the number of hypotheses $H\!=\!1$ and iterations $M\!=\!1$ during training, and $H\!=\!20$ and $M\!=\!10$ during inference, as in D3DP~\cite{Shan_2023_ICCV}.

\begin{table}[t]
\centering
\adjustbox{width=\linewidth}
 {
  \setlength{\tabcolsep}{19.5pt}
    \begin{tabular}{lcc}
    \toprule
    \multirow{2}{*}{Method} 
    & \multicolumn{2}{c}{Human3.6M (DET)} \\ 
    \cmidrule(lr){2-3}
    & MPJPE~$\textcolor{blue}{\downarrow}$ & P-MPJPE~$\textcolor{blue}{\downarrow}$ \\
    \midrule
    w/o Prompt  & 37.2 & 29.1\\
    M-Prompt & 35.8 & 28.1\\ 
    S-Prompt & 36.2 & 28.9\\  
    C-Prompt & 34.7 & 27.4\\  
    AL-Prompt & 34.6 & 27.4\\ 
    \midrule
    \rowcolor{Gray}\textbf{FinePOSE} (Ours) & \textbf{31.9} & \textbf{25.0} \\   
    \bottomrule
    \end{tabular}
}
\vspace{-7pt}
\caption{\small \textbf{Ablation study on different designs of prompt learning in the FPP block.} w/o Prompt: without any textual information and learnable prompts. M-Prompt: using the action class to design the prompt manually. S-Prompt: using a learnable prompt combined with the action class. C-Prompt: employing the action class and coarse-grained information to create the prompt. AL-Prompt: only learnable prompts without any manual design.}
\vspace{-12pt}
\label{tab:ablation-prompt}
\end{table}

\subsection{Comparison with the State-of-the-Arts}

\textbf{Human3.6M.} \cref{tab:SOTA-H3.6M} reports comparisons between our FinePOSE with state-of-the-art (SOTA) 3D HPE methods on the Human3.6M dataset. FinePOSE significantly achieves new SOTA performance, especially when using detected 2D keypoints as inputs.
Compared with existing 3D HPE methods, FinePOSE surpasses the SOTA method D3DP~\cite{Shan_2023_ICCV} by 3.5mm in MPJPE and 3.7mm in P-MPJPE.
When using ground truth 2D keypoints as inputs, FinePOSE also significantly outperforms the SOTA method MotionBERT~\cite{zhu2023motionbert}, improving MPJPE by 0.2mm.
\cref{tab:quantitive-H3.6M-det} provides detailed comparisons between on each action class using 2D keypoint detectors as inputs. For example, our FinePOSE achieves noticeable improvements (43.7mm$\rightarrow$36.5mm) for the action class ``\textit{Photo}'' and decreases average MPJPE by 3.5mm (35.4mm$\rightarrow$31.9mm).

\noindent\textbf{MPI-INF-3DHP.} \cref{tab:SOTA-3DHP} reports comparisons between our FinePOSE and SOTA 3D HPE methods on the MPI-INF-3DHP dataset, using ground truth 2D keypoints as inputs.
Compared with the SOTA existing method GLA-GCN~\cite{yu2023gla}, FinePOSE decreases MPJPE by 1.6mm and increases the PCK by 0.4\% and AUC by 0.9\%.
Overall, these experimental results demonstrate that our FinePOSE benefits from fine-grained part-aware prompt learning and pose-prompt communications, resulting in higher denoising quality and estimation accuracy.

\begin{table}[t]
\centering
\adjustbox{width=\linewidth}
 {
  \setlength{\tabcolsep}{6pt}
    \begin{tabular}{lccccc}
    \toprule
    \multirow{2}{*}{Method} 
    & \multicolumn{3}{c}{Configuration} 
    & \multirow{2}{*}{MPJPE~$\textcolor{blue}{\downarrow}$} & \multirow{2}{*}{P-MPJPE~$\textcolor{blue}{\downarrow}$} \\
    \cmidrule(lr){2-4}
    & FPP & FPC & PTS & & \\
    \midrule
    Baseline  & & & & 37.2 & 29.1\\
    w FPP     & \checkmark & & & 35.3 & 28.0 \\
    w/o FPP   & & & \checkmark & 37.1 & 29.2 \\
    w/o FPC   & \checkmark & & \checkmark &  35.7 & 27.8 \\
    w/o PTS   & \checkmark & \checkmark & & 36.6 & 29.0 \\
    \midrule
    \rowcolor{Gray}\textbf{FinePOSE} (Ours) & \checkmark & \checkmark & \checkmark & \textbf{31.9} & \textbf{25.0}\\  
    \bottomrule
    \end{tabular}
}
\vspace{-7pt}
\caption{\small \textbf{Ablation study on different configurations of FinePOSE on Human3.6M using 2D keypoint detectors as inputs.} Baseline: the method without any textual information via prompt learning. w FPP: the method only contains the FPP block and adds $\mathbf{P}[L]$ to the input. w/o FPP: the method without the FPP block leads to an infeasible FPC block. w/o FPC: the method without the FPC block. w/o PTS: the method without the PTS block.}
\vspace{-12pt}
\label{tab:ablation-block}
\end{table}

\subsection{Ablation Study}
We conduct a series of analysis experiments of our FinePOSE on the Human3.6M dataset to investigate the effects on the performance of different prompt learning designs in the FPP block and different blocks in FinePOSE. 

\begin{figure*}[t]
  \centering
  \includegraphics[width=\linewidth]{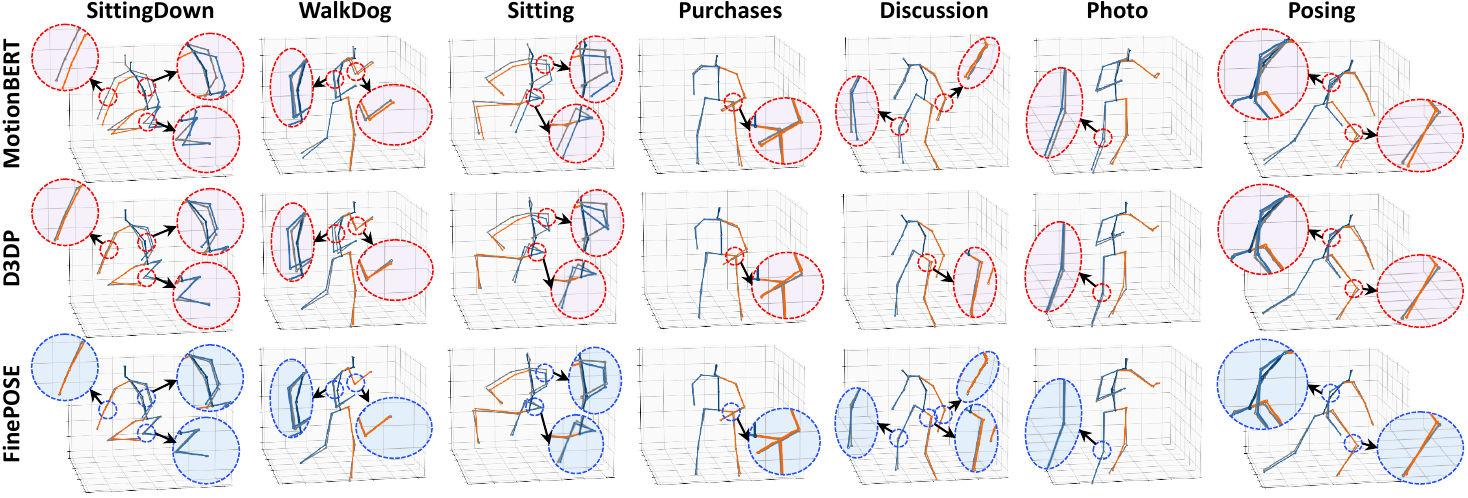}
    \vspace{-19pt}
  \caption{\textbf{Qualitative comparisons of our FinePOSE with MotionBERT~\cite{zhu2023motionbert} and D3DP~\cite{Shan_2023_ICCV} on Human3.6M.} The gray skeleton is the ground-truth 3D pose. The blue skeleton represents the prediction of the human left part, and the orange indicates the human right part. The red dashed line represents the incorrect regions of the compared methods, and the blue dashed line indicates the counterparts of FinePOSE.}
  \label{visualization}
  \vspace{-10pt}
\end{figure*}

\noindent\textbf{Effects of Different Designs in FPP.}
We design various versions of the FPP block for our FinePOSE, including a) w/o Prompt, b) M-Prompt, c) S-Prompt, d) C-Prompt, and e) AL-Prompt.
Specifically, w/o Prompt denotes FinePOSE without introducing textual information and learnable prompts. M-Prompt indicates using the action class to design the prompt manually instead of the FPP block. Taking the action class ``\textit{Directions}'' as an example, the manually designed prompt is ``\textit{a person is pointing directions with hands}''. There are 15 action classes available in the Human3.6M dataset corresponding to 15 kinds of manually designed prompts. S-Prompt indicates utilizing learnable prompts combined with the action class. C-Prompt indicates employing the action class and coarse-grained information like ``\textit{person}'' and ``\textit{speed}'' to create the prompt. Finally, AL-Prompt means only using learnable prompts without any manual design.

We first evaluate the effect of manually designed prompts (i.e., M-Prompt) on Human3.6M. As shown in \cref{tab:ablation-prompt}, compared to w/o Prompt, M-Prompt achieves a decrease of 1.4mm on MPJPE and 1.0mm on P-MPJPE, indicating that manually designing prompts is a practical strategy even though they cannot guarantee the prompt is optimal during the denoising process for the 3D HPE task.
To evaluate the effectiveness of S-Prompt, we compare it with w/o Prompt. As shown in \cref{tab:ablation-prompt}, MPJPE and P-MPJPE are reduced by 1.0mm and 0.2mm, respectively, for S-Prompt, which demonstrates that with the help of learnable prompts, integrating textual information can improve the performance on 3D HPE task. While compared to M-Prompt, S-Prompt results in performance degradation, indicating that learnable prompts must be meticulously designed. In addition, we also investigate the impact of manual intervention degrees on 3D HPE performance using two groups of comparative experiments. In the first group, we used only learnable prompts without any textual information and manual intervention, named AL-Prompt, which differs from S-Prompt with the action class. The second group designed a coarse-grained prompt involving action class, ``person'', ``speed'', and corresponding learnable prompts, denoted as C-Prompt. We see that both AL-Prompt and C-Prompt outperform S-Prompt since AL-Prompt is without interference from uncomplete textual information and C-Prompt contains some important textual information like action class, ``person'', and ``speed'', which provide the action subject and kinematic data.
Finally, it is observed that our FinePOSE outperforms various versions of prompt learning on both MPJPE and P-MPJPE, indicating the effectiveness of the fine-grained part-aware prompt learning mechanism in FinePOSE.

\begin{table}[t]
\centering
\adjustbox{width=\linewidth}
 {
  \setlength{\tabcolsep}{12pt}
    \begin{tabular}{lcccc}
    \toprule
    \multirow{2}{*}{Method / MPJPE~$\textcolor{blue}{\downarrow}$} 
    & \multicolumn{4}{c}{EgoHumans} \\ 
    \cmidrule(lr){2-5}
    & Tag. & Lego & Fenc. & Avg \\
    \midrule
    D3DP~\cite{d3dp-shan2023diffusion-iccv} & 30.7 & 29.0 & 46.6 & 35.4 \\
    \rowcolor{Gray}\textbf{FinePOSE} (Ours) & \textbf{30.0} & \textbf{26.7} & \textbf{46.2} & \textbf{34.3} \\   
    \rowcolor{Gray} & \textcolor{blue}{(-0.7)} & \textcolor{blue}{(-2.3)} & \textcolor{blue}{(-0.4)} & \textcolor{blue}{(-1.1)} \\
    \bottomrule
    \end{tabular}
}
\vspace{-7pt}
\caption{\textbf{Quantitative comparison with D3DP on the EgoHumans dataset using 2D keypoints as inputs.} \textit{Tag.}, \textit{Lego}, and \textit{Fenc.} correspond to 3 action classes. \textit{Avg} indicates the average MPJPE among 3 action classes.}
\vspace{-13pt}
\label{tab:multi_result}
\end{table}

\noindent\textbf{Effects of Different Blocks in FinePOSE.}
In \cref{tab:ablation-block}, we provide different settings of our FinePOSE to evaluate the effects of different blocks for the 3D HPE performance, including Baseline, w FPP, w/o FPP, w/o FPC, and w/o PTS. Specifically, Baseline denotes FinePOSE without introducing textual information and learnable prompts, the same as the configuration of w/o Prompt. w FPP indicates FinePOSE only contains the FPP block without introducing the FPC and PTS blocks and only adds textual information $\mathbf{P}[L]$ to the input. w/o FPP denotes FinePOSE without the FPP block, leading to the FPC block being infeasible and only utilizing the PTS block. w/o FPC means FinePOSE without the FPC block but using the FPP and PTS blocks. w/o PTS refers to FinePOSE without the PTS block but using the FPP and FPC blocks to integrate textual information for fine-grained part-aware prompt learning.

Compared w FPP and Baseline, we observe that the former can achieve 1.9mm and 1.1mm improvements on MPJPE and P-MPJPE. This is because our FinePOSE contains the FPP block, which adds the prompt embedding $\mathbf{P}[L]$ into the input $\mathbf{Z}_t$ of denoiser $\mathcal{D}$, significantly improving the denoising capability.
We observe that the results between w/o FPP and Baseline are almost equivalent. The baseline has already brought timestamp $t$ into the denoising process, while the PTS block refines the prediction at each noise level by reusing the timestamp to the denoising process after the FPP and FPC block. Thus, there is nearly no effect in adding only the PTS block without FPP and FPC blocks to the denoiser.
Making a comparison between w/o FPC and w/o FPP, the former achieves a decrease of 1.4mm on both MPJPE and P-MPJPE over w/o FPP, indicating that the FPP block in the denoiser plays a critical role in the fine-grained part-aware prompt learning mechanism.
Finally, we observe that FinePOSE achieves a decrease of 4.7mm on MPJPE and 4.0mm on P-MPJPE compared to w/o PTS, indicating the necessity to integrate learned prompt embeddings and timestamps in the PTS block.

\subsection{Results on 3D Multi-Human Pose Estimation}
In real-world applications, the multi-human scenario is more common than the single-human one. However, its complexity hinders existing work from handling it. In \cref{subsec:multipp}, we present a post-integration to extend FinePOSE for the multi-human pose estimation task. We implemented the extension using the SOTA method D3DP for a convincing comparison. The experimental results on EgoHumans are reported in \cref{tab:multi_result}, demonstrating that (1) the integration strategy indeed has potential feasibility and (2) FinePOSE has a dominant performance even in the complex multi-human scenario.

\subsection{Visualization}

\cref{visualization} shows the visualization results of D3DP~\cite{d3dp-shan2023diffusion-iccv}, MotionBERT~\cite{zhu2023motionbert} and our FinePOSE on Human3.6M. These methods have performed well for actions in which the body, legs, and other parts of the person in the scene are relatively clear. For the actions with simple shapes, e.g., ``Discussion'' and ``Photo'', the 3D poses predicted by FinePOSE match better with ground-truth 3D poses than those of D3DP and MotionBERT, especially in the left knee, right arm, and right hip of ``Discussion'' and in the left knee of ``Photo''. For the actions with complex shapes, e.g., ``Sitting'' and ``SittingDown'', FinePOSE is more accurate at various joints, especially for arms and legs, while the 3D poses predicted by D3DP and MotionBERT differ significantly from ground-truth 3D poses.

%% file: sec/5_conclusion.tex
\section{Conclusion and Discussion}
This work has presented FinePOSE, a new fine-grained prompt-driven denoiser for 3D human pose estimation. FinePOSE was composed of FPP, FPC, and PTS blocks. FPP learned fine-grained part-aware prompts to provide precise guidance for each human body part. FPC established fine-grained communication between learnable part-aware prompts and poses to enhance denoising capability. PTS brought timestamp information to the denoising process, strengthening the ability to refine the prediction at each noise level. Experimental results on two benchmarks demonstrated that FinePOSE surpasses the state-of-the-art methods. We have also extended FinePOSE from single-human scenarios to multi-human ones, exhibiting that our model performs well in complex multi-human scenarios.

\noindent\textbf{Limitations.} FinePOSE is not designed explicitly for the multi-person scenario. The diffusion model-based 3D HPE method is relatively computationally expensive.